\documentclass[a4paper,conference]{IEEEtran}
\usepackage{graphicx}
\usepackage{booktabs}
\usepackage{comment}
\usepackage{color}
\usepackage{hyperref}
\usepackage{tikz}
\usepackage{pgfplots}
\usetikzlibrary{arrows}

\ifCLASSINFOpdf
\else
\fi
\usepackage{amsmath, amssymb}
\usepackage{array, multirow}
\begin{document}
%
\title{
Graph-based Interpolation of Feature Vectors for Accurate Few-Shot Classification}

\author{\IEEEauthorblockN{Yuqing Hu}
\IEEEauthorblockA{Electronics Dept., IMT Atlantique\\
Orange Labs\\
France\\
Email: yuqing.hu@imt-atlantique.fr}
\and
\IEEEauthorblockN{Vincent Gripon}
\IEEEauthorblockA{Electronics Dept., IMT Atlantique\\
Brest, France\\
Email: vincent.gripon@imt-atlantique.fr}
\and
\IEEEauthorblockN{St\'ephane Pateux}
\IEEEauthorblockA{Orange Labs\\
Cesson-S\'evign\'e, France\\
Email: stephane.pateux@orange.com\\
}}


%


\maketitle

\begin{abstract}
In few-shot classification, the aim is to learn models able to discriminate classes using only a small number of labeled examples. In this context, works have proposed to introduce Graph Neural Networks (GNNs) aiming at exploiting the information contained in other samples treated concurrently, what is commonly referred to as the transductive setting in the literature. These GNNs are trained all together with a backbone feature extractor. In this paper, we propose a new method that relies on graphs only to interpolate feature vectors instead, resulting in a transductive learning setting with no additional parameters to train. Our proposed method thus exploits two levels of information: a) transfer features obtained on generic datasets, b) transductive information obtained from other samples to be classified. Using standard few-shot vision classification datasets, we demonstrate its ability to bring significant gains compared to other works.
\end{abstract}


%
\IEEEpeerreviewmaketitle

\section{Introduction}
\label{section:introduction}

Deep learning is the state-of-the-art solution for many problems in machine learning, specifically in the domain of computer vision. Relying on a huge number of tunable parameters, these systems are able to absorb subtle dependencies in the distribution of data in such a way that it can later generalize to unseen inputs. Numerous experiments in the field of vision suggest that there is a trade-off between the size of the model (for example expressed as the number of parameters~\cite{tan2019efficientnet}) and its performance on the considered task. As such, reaching state-of-the-art performance often requires to deploy complex architectures. On the other hand, using large models in the case of data-thrifty settings would lead to a case of an underdetermined system. This is why few-shot learning is particularly challenging in the field. 

In order to overcome this limitation of deep learning models, several works propose to use Graph Neural Networks (GNNs)~\cite{garcia2017few,kim2019edge,gidaris2019generating,liu2018learning}. GNNs are a natural way to exploit information available in other samples to classify, a setting often referred to as transductive in the literature. However, most often introduced GNNs come with their own set of parameters to be added to the already numerous parameters to tune to solve the considered task.  As a consequence, many of these methods do not achieve top-tier results when compared to state-of-the-art solutions.

\begin{figure*}
    \centering
    \begin{tikzpicture}[thick]
    \draw[fill=black,fill opacity=0.1,draw=black,draw opacity=0.1]
    (0,0) rectangle (4,4)
    (4.2,0) rectangle (8.2,4)
    (8.4,0) rectangle (12.4,4)
    (12.6,0) rectangle (17.,1.9)
    (12.6,2) rectangle (17.,4);
    \draw[]
    (-0.05,-0.05) rectangle (4.05,4.5)
    (4.15,-0.05) rectangle (12.45,4.5)
    (12.55,-0.05) rectangle (17.05,4.5)
    ;
    \node at (2,4.25) {\textbf{Pretraining}};
    \node at (8.4,4.25) {\textbf{Graph-based feature interpolation}};
    \node at (14.8,4.25) {\textbf{Logistic regression}};
    
    \begin{scope}[yshift=-0.2cm]
    \node at (2,3.8) {using lots of training data};
    \setlength{\fboxsep}{1pt}%
    \node at (1,3) {\fcolorbox{black}{white}{\includegraphics[width=1cm,height=1cm]{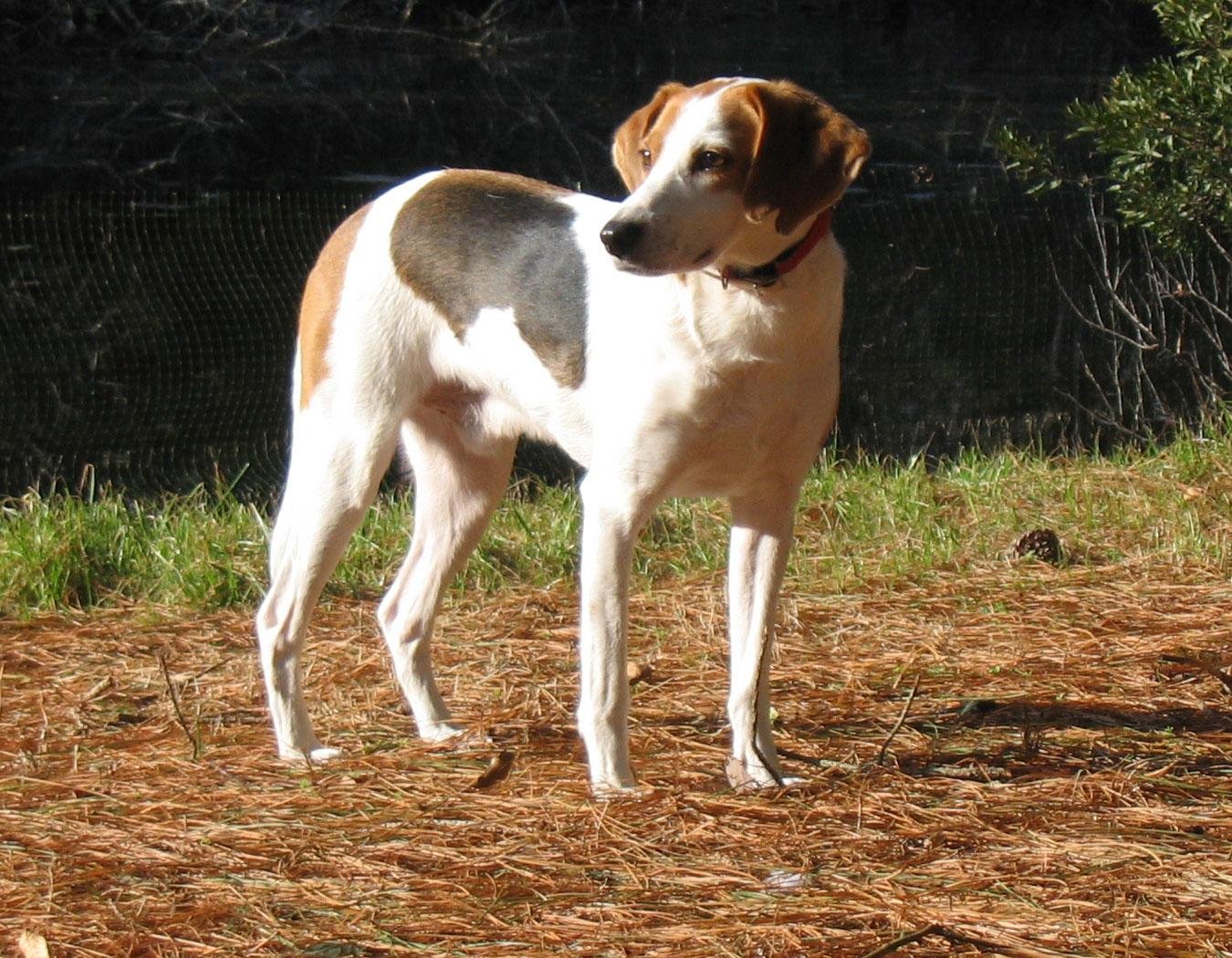}}};
    \node at (0.95,2.95) {\fcolorbox{black}{white}{\includegraphics[width=1cm,height=1cm]{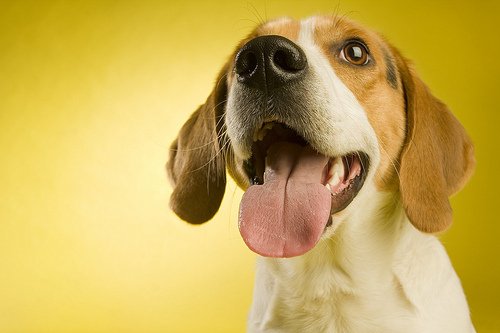}}};
    \node at (0.90,2.90) {\fcolorbox{black}{white}{\includegraphics[width=1cm,height=1cm]{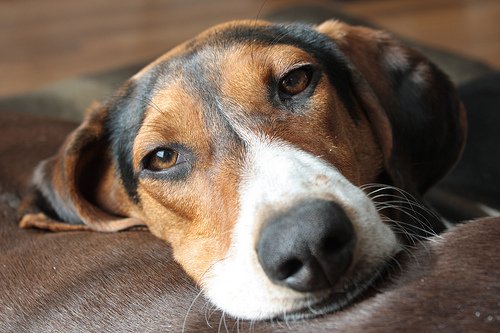}}};
    \node at (0.85,2.85) {\fcolorbox{black}{white}{\includegraphics[width=1cm,height=1cm]{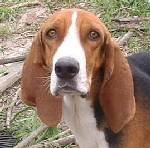}}};
    \node at (0.80,2.80) {\fcolorbox{black}{white}{\includegraphics[width=1cm,height=1cm]{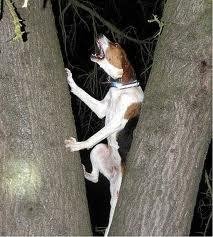}}};
    \node at (3,3) {\fcolorbox{black}{white}{\includegraphics[width=1cm,height=1cm]{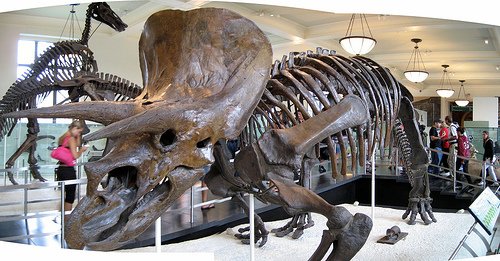}}};
    \node at (2.95,2.95) {\fcolorbox{black}{white}{\includegraphics[width=1cm,height=1cm]{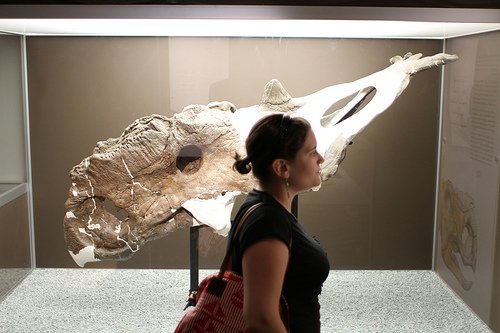}}};
    \node at (2.90,2.90) {\fcolorbox{black}{white}{\includegraphics[width=1cm,height=1cm]{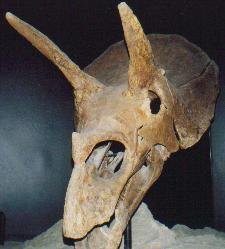}}};
    \node at (2.85,2.85) {\fcolorbox{black}{white}{\includegraphics[width=1cm,height=1cm]{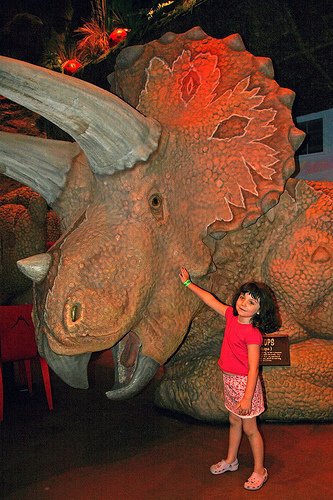}}};
    \node at (2.80,2.80) {\fcolorbox{black}{white}{\includegraphics[width=1cm,height=1cm]{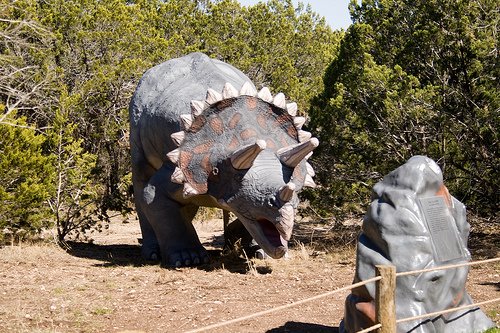}}};
    \node at (2,2.7) {\fcolorbox{black}{white}{\includegraphics[width=1cm,height=1cm]{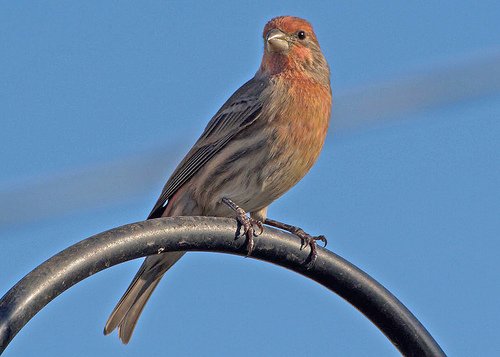}}};
    \node at (1.95,2.65) {\fcolorbox{black}{white}{\includegraphics[width=1cm,height=1cm]{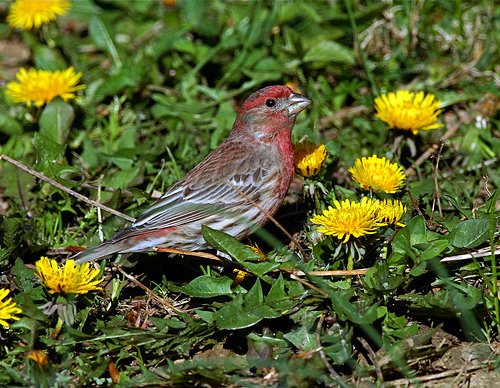}}};
    \node at (1.90,2.60) {\fcolorbox{black}{white}{\includegraphics[width=1cm,height=1cm]{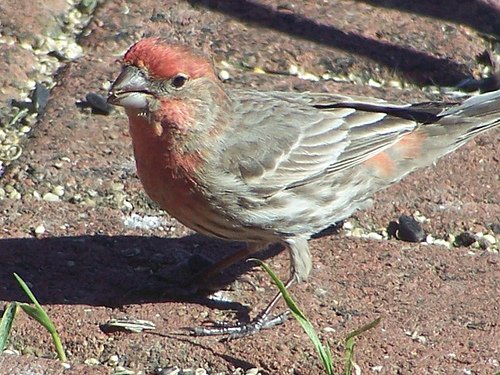}}};
    \node at (1.85,2.55) {\fcolorbox{black}{white}{\includegraphics[width=1cm,height=1cm]{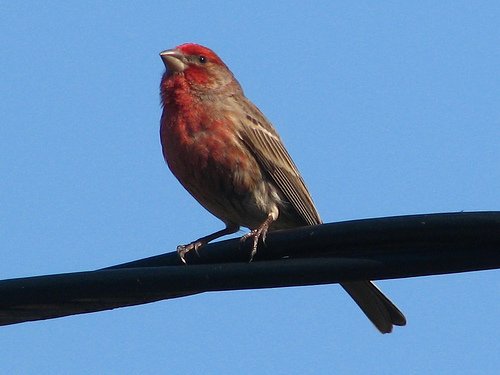}}};
    \node at (1.80,2.50) {\fcolorbox{black}{white}{\includegraphics[width=1cm,height=1cm]{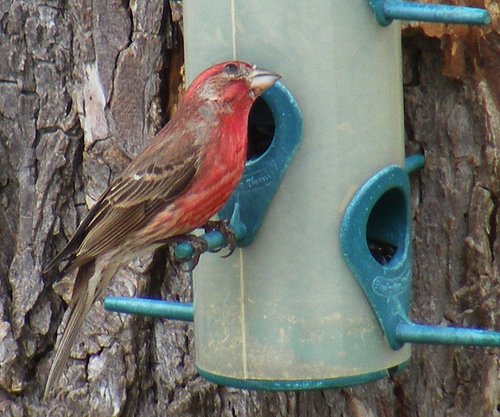}}};
    \end{scope}
    \node[blue] at (2,1.2) {1.train feature extractor};
    \node at (2,0.7) {$\mathbf{x}_i \mapsto f_\varphi(\mathbf{x}_i) = \mathbf{v}_i$};
    \node at (0.8,0.2) {\tiny{novel input}};
    \node at (3.15,0.2) {\tiny{feature vector}};
    \path[->,>=stealth]
    (3.1,0.3) edge (3.1,0.6)
    (0.85,0.3) edge (0.85,0.6);
    
    \begin{scope}[yshift=-0.1cm]
    \tikzstyle{every node} = [draw,circle,fill=white, scale=0.71];
    \node(v1) at (4.8, 3.3) {$\mathbf{v}_1$};
    \node(v2) at (7.5, 2.8) {$\mathbf{v}_2$};
    \node(v3) at (6.2,3.2) {$\mathbf{v}_3$};
    \node(v4) at (6.2,1.7) {$\mathbf{v}_4$};
    \node(v5) at (4.8,1.4) {$\mathbf{v}_5$};
    \node(v6) at (7.5,1.4) {$\mathbf{v}_6$};
    \tikzstyle{every node}=[scale=0.71];
    \path
    (v1) edge (v3)
    (v2) edge (v3)
    edge (v6)
    (v3) edge (v4)
    (v4) edge (v5)
    edge (v6);
    \path[->,>=stealth']
    (6.7,3.6) edge (6.85,3.0);
    \node at (6.8,3.8) {based on $\cos(\mathbf{v}_2,\mathbf{v}_3)$};
    \end{scope}
    \node[blue] at (6.2, 0.5) {2.construct similarity graph};

    \begin{scope}[xshift=4.2cm,yshift=0.3cm]
    \tikzstyle{every node} = [draw,circle,fill=white,minimum width=18pt,scale=0.71];
    \node(v1) at (4.8, 3.3) {};
    \node(v2) at (7.5, 2.8) {};
    \node(v3) at (6.2,3.2) {};
    \node(v4) at (6.2,1.7) {};
    \node(v5) at (4.8,1.4) {};
    \node(v6) at (7.5,1.4) {};
    \tikzstyle{every node}=[scale=0.71];
    \path
    (v1) edge (v3)
    (v2) edge (v3)
    edge (v6)
    (v3) edge (v4)
    (v4) edge (v5)
    edge (v6);
    \path[>=stealth',<->]
    (v1) edge (v3)
    (v2) edge (v3)
    edge (v6)
    (v3) edge (v4)
    (v4) edge (v5)
    edge (v6);
    \end{scope}
    \node at (10.4,1.) {$\mathbf{V}_{new}=(\alpha \mathbf{I} + \mathbf{E})^\kappa\mathbf{V}$};
    \node[blue] at (10.4,0.4) {3.feature propagation};


    \node[blue] at (14.8, 2.2) {\small{4.training with labeled inputs}};
    \tikzstyle{every node} = [draw,circle,minimum width=7pt,inner sep =0pt,scale=0.71];
    \begin{scope}[xshift=0.8cm,yshift=-0.1cm]
    \foreach \i in {0,...,6}{
        \node(a\i) at (13.+0.3*\i,3.5) {};
    }
    \foreach \i in {0,...,3}{
        \node(b\i) at (13.45+0.3*\i,2.7) {};
    }
    \path[opacity=0.8]
    \foreach \i in {0,...,6}{
        \foreach \j in {0,...,3}{
            (a\i) edge (b\j)
        }
    };
    \tikzstyle{every node}=[scale=0.65];
    \node at (14.,3.9) {\footnotesize{train mapping $\mathbf{V}_{new}^{\tiny{\text{labeled}}}$ to labels}};
    \end{scope}
    
    \tikzstyle{every node} = [draw,circle,minimum width=7pt,inner sep =0pt,scale=0.71];
    \begin{scope}[xshift=0.8cm,yshift=-2.2cm]
    \foreach \i in {0,...,6}{
        \node(a\i) at (13.+0.3*\i,3.5) {};
    }
    \foreach \i in {0,...,3}{
        \node(b\i) at (13.45+0.3*\i,2.7) {};
    }
    \path[opacity=0.8]
    \foreach \i in {0,...,6}{
        \foreach \j in {0,...,3}{
            (a\i) edge (b\j)
        }
    };
    \tikzstyle{every node}=[scale=0.62];
    \node at (14,3.9) {\footnotesize{predict $\mathbf{V}_{new}^{\text{\tiny{unlabeled}}}$ associated labels}};
    \end{scope}
    \tikzstyle{every node}=[scale=0.71];
    \node[blue] at (14.8, 0.15) {\small{5.prediction on other inputs}};
    
    \end{tikzpicture}
    \caption{Illustration of the proposed method. The proposed method is composed of three stages. During the pretraining stage, a classical backbone is trained using large datasets (step 1.). This trained backbone is then used to extract features of a novel dataset, comprising few supervised inputs. During feature interpolation, first is built a similarity graph depending on the cosine similarity between extracted features of both labeled and unlabeled available data (step 2.). Then this graph is used to diffuse (i.e. interpolate) features of similar (neighbor) samples (step 3.). The obtained representations are used to train a simple logistic classifier (step 4.) using the supervised data. Finally, in step 5., the trained classifier is used to perform predictions on unlabeled data.}
    \label{fig:illustration}
\end{figure*}
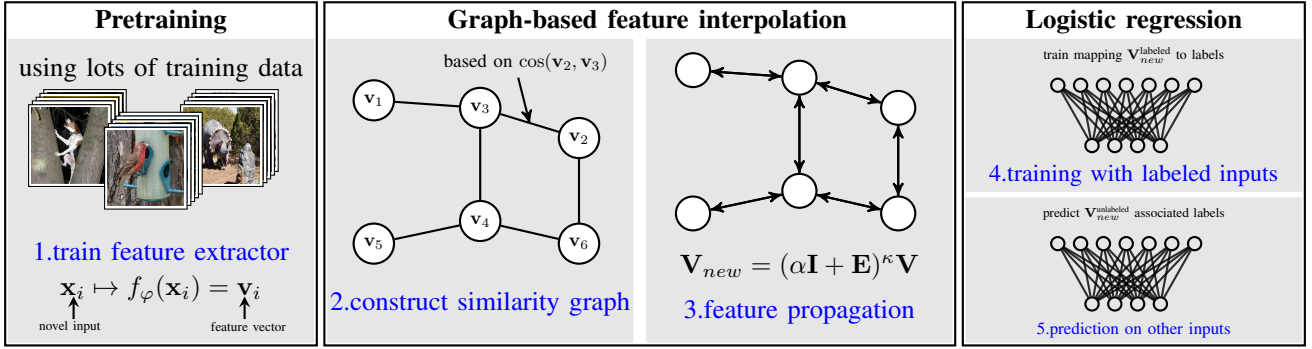

In this work, we propose to incorporate a graph-based method with no additional parameters, as a way to naturally bring transductive information in solving the considered task. The first step of the method consists in training a feature extractor with abundant data, followed by an interpolation strategy using well designed graphs. The graphs considered in this paper use vertices to represent each sample of the batch, and their edges are weighted depending on the similarity of corresponding feature vectors. The graph is thus used to interpolate features and thus share information between inputs. Once the features have been interpolated, we simply use a classical Logistic Regression (LR) to classify them. This work comes with the following claims:
\begin{itemize}
    \item We introduce a three-stage method for few-shot classification of input images that combines state-of-the-art transfer learning~\cite{mangla2020charting}, a graph-based interpolation technique and logistic regression.
    \item We empirically demonstrate that the proposed method reaches competitive accuracy on standardized benchmarks in the field of few-shot learning and largely surpasses the current works using GNNs.
    \item We analyze the importance of each step of the method and discuss hyperparameters influence.
\end{itemize}

The paper is organized as follows. In Section~\ref{sec:rw}, we present related works. In Section~\ref{sec:met} we introduce our proposed methodology. In Section~\ref{sec:exp}, we show experimental results on standard vision datasets and discuss hyperparameters influence. Finally, Section~\ref{sec:con} is a conclusion.
The source code can be found at~\url{https://github.com/yhu01/transfer-sgc}.

\section{Related Work}
\label{sec:rw}
\textbf{Optimization based methods:} Recent work on few-shot classification contains a variety of approaches, some of which can be categorized as meta-learning~\cite{thrun2012learning} where the goal is to train an optimizer that initializes the network parameters using a first generic dataset, so that the model is able to reach good performance with only a few more steps on actual considered data. The well-known MAML method~\cite{finn2017model} trains on different tasks with a basic stochastic gradient decent optimizer~\cite{chen2019closer} and Meta-LSTM~\cite{ravi2016optimization} utilizes a LSTM-based meta-learner that is thus memory-augmented. Meta-learning can be thought of as a refined transfer method, where the few-shot setting is taken into consideration directly when training on the generic dataset. Although both MAML and Meta-LSTM achieve good performance with quick adaptation, this type of solution suffers from the domain shift problem~\cite{chen2019closer} as well as the sensitivity of hyperparameters.

\textbf{Embedding based methods:} Another popular approach aims at finding compact embedding for the input data by learning a metric that measures the distance in a low-dimensional way. Matching Nets~\cite{vinyals2016matching} and Proto Nets~\cite{snell2017prototypical} learn a nearest-neighbor classifier by comparing the distance between the query inputs and labeled inputs with a certain metric, while Relation Nets~\cite{sung2018learning} construct a new neural network that learns the metric itself. If some of these methods are able to outperform MAML, they mainly suffer from over-fitting and a lack of task specific information. 

Therefore, ideas have been proposed to address these issues. For example in~\cite{li2019finding}, a plug network is added to find task-relevant features inside embeddings so that the model can tell the inter-class uniqueness and intra-class commonality for a specific task. In~\cite{lee2019meta} and~\cite{bertinetto2018meta}, the authors create a class-weight generator by training the model with a linear classifier (e.g. SVM) in order for the model to minimize generalization error across a distribution of tasks. More recently, the use of graph methods~\cite{gori2005new}~\cite{koch2015siamese} starts to gain momentum in the few-shot learning problems. For example, in~\cite{garcia2017few,kim2019edge,gidaris2019generating,liu2018learning}, the authors incorporate the idea of semi-supervised learning~\cite{chapelle2009semi} as a mean to benefit from the unlabeled query input data when solving a task, what is referred to as the transductive setting. Many recent works propose neural networks able to handle inputs supported on graphs~\cite{hamilton2017representation}. For example, in GCN~\cite{kipf2016semi}, the authors introduce a graph convolution operator, that can be used in cascade to generate deep learning architectures. In GAT~\cite{velivckovic2017graph}, the authors enrich GCN with additional learnable attention kernels. In SGC~\cite{wu2019simplifying}, the authors propose to simplify GCN by using only one-layer systems on powers of the adjacency matrix of considered graphs. Interestingly, they reach state-of-the-art accuracy with fewer parameters.

\textbf{Hallucination based methods:} Other methods propose to augment the training sets by learning a generator that can hallucinate novel class data using data-augmentation techniques~\cite{chen2019closer}. In~\cite{zhang2019few}, the authors extract labeled data into different components and then combine them using learned transformations, while in~\cite{chen2019image}, the authors aim at constructively deforming original samples with new samples drawn from another dataset. However, these methods lack precision as in the way the data is generated, which results in coarse and low-quality synthesized data that can sometimes lead to unsignificant gains in performance~\cite{wang2019few}. 

\textbf{Transfer based methods:} As in our work, transfer learning is another possible solution to solve few-shot classification problems. The main idea is to first train a feature extractor using a generic dataset~\cite{torrey2010transfer,das2019two}, then process these features directly when solving the new task. In~\cite{chen2019closer} a distance-based classifier is applied to train the backbone (i.e. the feature extractor), and in~\cite{mangla2020charting}, the authors aims at improving the feature quality by adding self-supervised learning and data-augmentation techniques during training. These methods have been proven to perform generally well, yet the challenge remains to fine-tune using the limited amount of labeled data.

In our work, we propose to align multiple ingredients that have been introduced in this section. Namely, we use transfer with graph-based interpolation. We mainly use transfer to exploit information contained in massive generic datasets, and we use a graph method to leverage the additional information available in both labeled and unlabeled inputs. Following the transductive setting, our proposed method can be considered as similar to~\cite{liu2018learning,garcia2017few,kim2019edge,gidaris2019generating}, but contrary to their works, we adopt a strategy in which the considered graph-based method contains no additional parameters to be trained. Our method can also be seen as a modification of Simplified Graph Convolutions~\cite{wu2019simplifying}, where contrary to their work we infer a graph structure from the latent representations of data.

\section{Methodology}
\label{sec:met}

\subsection{Problem statement}
Consider the following problem. We are given two datasets, termed $\mathbf{D}_{base}$ and $\mathbf{D}_{novel}$ with disjoint classes. The first one (called ``base'') contains a large number of labeled examples from $K_b$ different classes. The second one (called ``novel'') contains a small number of labeled examples, along with some unlabeled ones, all from $K_n$ new classes. Our aim is to accurately predict the class of the unlabeled inputs of the novel dataset. There are a few important parameters to this problem: the number of classes in the novel dataset $K_n$, the number of training samples $s$ for each corresponding class, and the total number of unlabeled inputs $Q$.

Note that in previous works~\cite{liu2018learning}, authors consider that there are exactly $q = Q/K_n$ unlabeled inputs for each class. We consider that this is non-practical, since in most applications there is no reason to think that this holds. We shall see in Section~\ref{sec:exp} that this has strong implications in terms of performance, especially when $q$ is small. Indeed, in practice the $Q$ unlabeled examples are drawn uniformly at random in a pool containing the same amount of unlabeled inputs for each class. So, when $Q$ is large, the central limit theorem tells us that the number of drawn inputs from each class should be similar, whereas it can be highly contrasted when $Q$ is small, leading to an imbalanced case.

\subsection{Proposed solution}

Our method is illustrated in Figure~\ref{fig:illustration}.
We first train a backbone deep neural network able to discriminate inputs from the base dataset $\mathbf{D}_{base}=\{(\mathbf{x'}_1, \ell_1),..., (\mathbf{x'}_m, \ell_m)\}$, where $\mathbf{x'}_i\in\mathbb{R}^d$ and $1\leq \ell_i \leq K_b$. The proposed methodology builds upon using this pretrained architecture as a generic feature extractor, what is referred to as \emph{transfer} in the literature~\cite{torrey2010transfer}. Usually, a common way to extract features is to process data belonging to the novel dataset using the penultimate activation layer. Here, we obtain the extractor $f_\varphi:\mathbb{R}^d\rightarrow\mathbb{R}^h$, where $\varphi$ are the learnable parameters trained using only the base dataset.

We then directly make use of the transferred representations $f_\varphi(\mathbf{D}_{novel}) = \{f_\varphi(\mathbf{x}), \mathbf{x}\in \mathbf{D}_{novel}\}$. Based on these, we build a $k$ nearest neighbor graph using cosine similarity: $$\cos(f_\varphi(\mathbf{x}),f_\varphi(\mathbf{y})) = \frac{f_\varphi(\mathbf{x})^\top f_\varphi(\mathbf{y})}{\|f_\varphi(\mathbf{x})\|_2\|f_\varphi(\mathbf{y})\|_2}.$$
This graph contains as many vertices as the total number of inputs in the novel dataset (both labeled and unlabeled ones).
Then, we train a model of simplified graph convolution model, that is supervised only for labeled inputs.

The rationale behind this method is twofold: $1)$ the pretrained backbone should be able to find good discriminative features since it is trained on a sufficiently large labeled dataset $2)$ the graph-based interpolation technique should be able to benefit from both the supervised inputs and the unlabeled ones, resulting in significant gains in accuracy when compared to methods that would ignore the unlabeled data. 

We show in the experiments that this method is also able to outperform other methods that use the unlabeled data especially when the number of labeled inputs is very limited.

The details of the proposed method are provided in the following paragraphs, first the pre-training stage (i.e. training the generic backbone), followed by the feature interpolation and logistic regression stages.

\textbf{Pre-training:} We follow the methodology introduced in~\cite{mangla2020charting}. In more details the feature extractor $f_\varphi$ and a distance-based classifier $D_{\mathbf{W}_b}$(parametrized by $\mathbf{W}_b$)~\cite{mensink2012metric} are trained on $\mathbf{D}_{base}$, where we compute the cosine distance between an input feature $f_\varphi(\mathbf{x'}_i)$ and each weight vector in $\mathbf{W}_b$ in order to reduce the intra-class variations~\cite{chen2019closer}. The training process consists of two sub-stages: the first sub-stage utilizes rotation-based self-supervised learning technique~\cite{gidaris2018unsupervised} where each input image is randomly rotated by a multiple of 90 degrees. We then co-train a linear classifier to tell which rotation was applied. Therefore, the total loss function of this sub-stage is given by:
\begin{equation}
L_A=L_\text{class}+L_\text{rotation}.
\label{eq:loss1}
\end{equation}
The second sub-stage fine-tunes the model with Manifold Mixup~\cite{verma2018manifold} technique for a few more epochs, where the outputs of hidden layers in the neural network are linearly combined to help the trained model generalize better. The total loss in this sub-stage is given by:
\begin{equation}
L_B=L_\text{ManifoldMixup} + 0.5(L_\text{class}+L_\text{rotation}).
\label{eq:loss2}
\end{equation}
With this training process, we are able to obtain robust input representations that generalize well to novel classes.

\textbf{Feature interpolation:} We consider fixed the pretrained parameters $\varphi$ of $f_\varphi$. Before training a new classifier $C_{\mathbf{W}_n}$ on the transferred representations of the novel dataset, we propose to interpolate features using a graph.

In details, we define a graph $G_T(\mathbf{V}, \textbf{E})$~\cite{kipf2016semi} where vertices matrix $\mathbf{V}\in\mathbb{R}^{(s K_n+Q)\times h}$ contains the stacked features of labeled and unlabeled inputs~\cite{garcia2017few}. To build the adjacency matrix $\mathbf{E}\in\mathbb{R}^{(s K_n+Q)\times (s K_n+Q)}$, we first compute:
\begin{equation}
\mathbf{S}[i,j]=\left\{\begin{array}{ll}\cos(\mathbf{V}[i,:],\mathbf{V}[j,:])& \text{$if $} i \neq j\\0 & \text{$otherwise$}\end{array}\right.,
\label{eq:simfunc}
\end{equation}
where $\mathbf{V}[i,:]$ denotes the $i$-th row of $\mathbf{V}$. Note that in all backbone architectures we use in the experiments, the penultimate layers are obtained by applying a ReLU function, so that all coefficients in $\mathbf{V}$ are nonnegative. As a result, coefficients in $\mathbf{S}$ are nonnegative as well. Also, note that $\mathbf{S}$ is symmetric.

Then, we only keep the value $\mathbf{S}[i,j]$ if it is one of the $k$ largest values on the corresponding row or on the corresponding column in $\mathbf{S}$. So, as soon as $k\geq (s K_n + Q - 1)$, all values are kept. Otherwise, $\mathbf{S}$ contains many 0s.

Finally, we apply normalization on the resulting matrix: 
\begin{equation}
\mathbf{E} = \mathbf{D}^{-1/2}\mathbf{S}\mathbf{D}^{-1/2},
\end{equation}
where $\mathbf{D}$ is the degree diagonal matrix defined as: $$\mathbf{D}[i,i] = \sum_j{\mathbf{S}[i,j]}.$$
Therefore, the graph vertices represent all inputs (both labeled and unlabeled) of the novel dataset. Its nonzero weights are based on the cosine similarity between corresponding transferred representations.

We then apply feature propagation~\cite{wu2019simplifying} to obtain new features for each vertex. The formula is:
\begin{equation}
\mathbf{V}_{new}=\underbrace{(\alpha \mathbf{I} + \mathbf{E})^\kappa}_{\text{``diffusion matrix''}}\mathbf{V},
\label{eq:featurepro}
\end{equation}
in which $\kappa$ and $\alpha$ are both hyperparameters, and $\mathbf{I}$ is the identity matrix. The role of $\kappa$ is important: providing $\kappa$ is too small, the new feature of a vertex will only depend on its direct neighbors in the graph. Using larger values of $\kappa$ allows to encompass for more indirect relationships. Using a too large value of $\kappa$ might drown out the information by averaging over all inputs. Similarly, $\alpha$ allows to balance between the neighbors representations and self-ones.

\textbf{Logistic regression:} Finally, a softmax classifier is trained using only the labeled vertices. We denote by $\mathbf{V}_{new}^{\text{labeled}}$ the subset of $\mathbf{V}_{new}$ corresponding to labeled vertices, then the predicted results $\hat{\mathbf{Y}}$ can be written following this formula:
\begin{equation}
\hat{\mathbf{Y}}^{\text{labeled}}=softmax(\mathbf{V}_{new}^{\text{labeled}}\mathbf{W}_n),
\label{eq:yhat}
\end{equation}
where $\mathbf{V}_{new}^{\text{labeled}}\in\mathbb{R}^{(s K_n)\times h}$, $\hat{\mathbf{Y}}\in\mathbb{R}^{(s K_n)\times K_n}$ and $\hat{\mathbf{Y}}[i,j]$ denotes the probability of vertex $i$ being categorized as being in the $j$-th class. 

Prediction is performed using the same principle, but using unlabeled inputs instead: denote by $\mathbf{V}_{new}^{\text{unlabeled}}$ the subset of $\mathbf{V}_{new}$ corresponding to unlabeled inputs, then we have the decision:
\begin{equation}
    \hat{\mathbf{Y}}^{\text{unlabeled}}[i] = \arg\max_j((\mathbf{V}_{new}^{\text{unlabeled}}\mathbf{W}_n)[i,j]).
\end{equation}

In Table~\ref{tab:hyperparams} we summarize the main parameters and hyperparameters of the considered problem and proposed solution. Let us point out that the proposed graph-based method does not contain any parameter to train.

\begin{table}[h]
    \caption{Parameters and hyperparameters of the considered problem and proposed solution (\# stands for ``number'').}
    \centering
    \scalebox{1.2}{
    \begin{tabular}{|c|l|}
    \hline
    \multicolumn{2}{|c|}{Novel dataset parameters}\\
    \hline
         $K_n$& \# classes\\
         \hline
         $s$& \# supervised inputs per class\\
         \hline
         $Q$ & total \# of unsupervised inputs\\
         \hline
         \hline

    \multicolumn{2}{|c|}{Proposed method hyperparameters}\\
\hline
         $1 \leq k < s K_n + Q$ & \# nearest neighbors to keep\\
         \hline
         $\kappa\in \mathbb{N}^*$ & power of the diffusion matrix\\
         \hline
         $0\leq \alpha \leq 1$ & strength of self-representations\\
         \hline
    \end{tabular}
    }
    \label{tab:hyperparams}
\end{table}

\section{Experimental Validation}
\label{sec:exp}
\subsection{Datasets}

We perform our experiments on 3 standardized few-shot classification datasets: miniImageNet~\cite{vinyals2016matching}, CUB~\cite{wah2011caltech} and CIFAR-FS~\cite{bertinetto2018meta}. These datasets are split into two parts: a) $K_b$ classes are chosen to train the backbone, called base classes, b) $K_n$ classes are drawn uniformly in the remaining classes to form the novel dataset, called novel classes. Among the $K_n$ drawn novel classes, $s$ labeled inputs per class and a total of $Q$ unlabeled inputs are drawn uniformly at random. As in most related works, unless mentioned otherwise all our experiments are performed using $K_n=5$ and $Q/K_n = 15$. We perform a run of 10,000 random draws to obtain an accuracy score and indicate confidence scores (95\%) when relevant.

\textbf{miniImageNet:} It consists of a subset of ImageNet~\cite{russakovsky2015imagenet} that contains 100 classes and 600 images of size $84\times84$ pixels per class. According to the standard~\cite{ravi2016optimization}, we use 64 base classes to train the backbone and 20 novel classes to draw the novel datasets from. So, for each run, 5 classes are drawn uniformly at random among these 20 classes.

\textbf{CUB:} The dataset contains 200 classes and has a total of 11,788 images of size $84\times84$ pixels. We split it into 100 base classes to train the backbone and 50 novel classes to draw the novel datasets from.

\textbf{CIFAR-FS:} This dataset has 100 classes, each class contains 600 images of size $32\times32$ pixels. We use the same numbers as for the miniImageNet dataset.

\subsection{Backbone models and implementation details}
We perform experiments using 2 different backbones as the structure of feature extractor $f_\varphi(\mathbf{x})$.

\textbf{Wide residual networks (WRN)}~\cite{zagoruyko2016wide}\textbf{:} We follow the settings in~\cite{mangla2020charting} by choosing a WRN with 28 convolutional layers and a widening factor of 10. The output feature size $h$ is 640.

\textbf{Residual networks (ResNet18)}~\cite{he2016deep}\textbf{:} Our ResNet18 contains a total of 18 convolutional layers grouped into 8 blocks. Following the settings in~\cite{wang2019simpleshot}, we remove the first two down-sampling layers and change the kernel size of the first convolutional layer to $3\times3$ pixels instead of $7\times7$ pixels. Here, $h=512$.

For the pre-training stage and miniImageNet, we train all backbones for a total of $470$ epochs from scratch using Adam optimizer~\cite{kingma2014adam} and cross-entropy loss, including $400$ epochs on the first sub-stage and $70$ epochs on the second sub-stage. For the logistic regression, we train with the same optimizer and loss function for $1000$ epochs with learning rate being $1e-3$ and weight decay being $5e-6$, which typically requires of the order of one second of computation on a modern GPU. Note that we observed that convergence usually occurs much quicker than 1000 epochs. In the In-Domain settings two stages are trained on the same dataset with base classes and novel classes respectively, while in the Cross-Domain settings we use these splits from two different datasets (e.g. base classes from miniImageNet and novel classes from CUB).

\subsection{Comparison with state-of-the-art methods}

As a first experiment, we compare the raw performance of the proposed method with state-of-the-art solutions with WRN and ResNet18 as backbones. The results are presented in Table~\ref{tab:results}. We fixed $\alpha$, $k$ and $\kappa$ respectfully with $s=1$ and $s=5$ for the proposed method, as it empirically gave the best results. Note that the sensitivity of these hyperparameters is discussed later in this section.

We point out that the proposed method reaches state-of-the-art performance in both case of 1-shot and 5-shot classification for most of the time, whatever the choice of all considered datasets. Note that the gain we observe is higher in the 1-shot case than in the 5-shot case, this is expected as in the case of 1-shot, the unlabeled samples bring proportionally more information compared to the case of 5-shot. In the extreme case of $s$-shot, with $s$ large enough, we expect the unlabeled samples to be almost useless.

We also perform experiments where the backbone has been trained using the base classes of miniImageNet but the few-shot task is performed using the novel classes of the CUB dataset. According to the results, we can draw conclusions very similar to the previous study, where the proposed method performs well for this specific task.

\begin{table*}
    \caption{1-shot and 5-shot accuracy of state-of-the-art methods in the literature, compared with the proposed solution. We present results using WRN and ResNet18 as backbones. For the proposed solution, we use the hyperparameters $\alpha=0.5$, $k=10$ and $\kappa=3$ for $s=1$; $\alpha=0.75$, $k=15$ and $\kappa=1$ for $s=5$.}
    \centering
    \scalebox{1.0}{
    \begin{tabular}{l|l|l|l}
         \toprule
         &          & \multicolumn{2}{c}{\textbf{miniImageNet}} \\
         Method & Backbone & 1-shot & 5-shot \\
         \midrule
         MAML~\cite{finn2017model} & ResNet18 & $49.61\pm0.92\%$ & $65.72\pm0.77\%$\\
         Baseline++~\cite{chen2019closer} & ResNet18 & $51.87\pm0.77\%$ & $75.68\pm0.63\%$\\
         Matching Networks~\cite{vinyals2016matching} & ResNet18 & $52.91\pm0.88\%$ & $68.88\pm0.69\%$\\ 
         ProtoNet~\cite{snell2017prototypical} & ResNet18 & $54.16\pm0.82\%$ & $73.68\pm0.65\%$\\
         SimpleShot~\cite{wang2019simpleshot} & ResNet18 & $63.10\pm0.20\%$ & $79.92\pm0.14\%$\\
         S2M2\_R~\cite{mangla2020charting} & ResNet18 & $64.06\pm0.18\%$ & $80.58\pm0.12\%$\\
         LaplacianShot~\cite{ziko2020laplacian} & ResNet18 & $72.11\pm0.19\%$ & $82.31\pm0.14\%$\\ 
         Transfer+Graph Interpolation (ours) & ResNet18 & $\mathbf{72.40\pm0.24}\%$ & $\mathbf{82.89\pm0.14}\%$\\
         \midrule
         ProtoNet~\cite{snell2017prototypical} & WRN & $62.60\pm0.20\%$ & $79.97\pm0.14\%$\\
         Matching Networks~\cite{vinyals2016matching} & WRN & $64.03\pm0.20\%$ & $76.32\pm0.16\%$\\
         S2M2\_R~\cite{mangla2020charting} & WRN & $64.93\pm0.18\%$ & $83.18\pm0.11\%$\\
         SimpleShot~\cite{wang2019simpleshot} & WRN & $65.87\pm0.20\%$ & $82.09\pm0.14\%$\\
         SIB~\cite{hu2020empirical} & WRN & $70.00\pm0.60\%$ & $79.20\pm0.40\%$\\
         BD-CSPN~\cite{liu2019prototype} & WRN & $70.31\pm0.93\%$ & $81.89\pm0.60\%$\\
         LaplacianShot~\cite{ziko2020laplacian} & WRN & $74.86\pm0.19\%$ & $84.13\pm0.14\%$\\
         Transfer+Graph Interpolation (ours) & WRN & $\mathbf{76.50\pm0.23}\%$ & $\mathbf{85.23\pm0.13}\%$\\
         \bottomrule
         
         
         \toprule
         &          & \multicolumn{2}{c}{\textbf{CUB}} \\
         Method & Backbone & 1-shot & 5-shot \\       
         \midrule
         S2M2\_R~\cite{mangla2020charting} & ResNet18 & $71.43\pm0.28\%$ & $85.55\pm0.52\%$\\
         ProtoNet~\cite{snell2017prototypical} & ResNet18 & $72.99\pm0.88\%$ & $86.64\pm0.51\%$\\
         Matching Networks~\cite{vinyals2016matching} & ResNet18 & $73.49\pm0.89\%$ & $84.45\pm0.58\%$\\ 
         LaplacianShot~\cite{ziko2020laplacian} & ResNet18 & $80.96\%$ & $88.68\%$\\
         Transfer+Graph Interpolation (ours) & ResNet18 & $\mathbf{86.05\pm0.20\%}$ & $\mathbf{90.87\pm0.10\%}$\\
         \midrule
         S2M2\_R~\cite{mangla2020charting} & WRN & $80.68\pm0.81\%$ & $90.85\pm0.44\%$\\
         Transfer+Graph Interpolation (ours) & WRN & $\mathbf{88.35\pm0.19\%}$ & $\mathbf{92.14\pm0.10\%}$\\
         \bottomrule
         
         \toprule
         &          & \multicolumn{2}{c}{\textbf{miniImageNet}$\longrightarrow$\textbf{CUB}} \\
         Method & Backbone & 1-shot & 5-shot \\
         \midrule
         Baseline++~\cite{chen2019closer} & ResNet18 & $40.44\pm0.75\%$ & $56.64\pm0.72\%$\\
         SimpleShot~\cite{wang2019simpleshot} & ResNet18 & $48.56\%$ & $65.63\%$\\
         LaplacianShot~\cite{ziko2020laplacian} & ResNet18 & $\mathbf{55.46}\%$ & $66.33\%$\\
         Transfer+Graph Interpolation (ours) & ResNet18 & $51.67\pm0.24\%$ & $\mathbf{69.83\pm0.18\%}$\\
         \midrule
         Manifold Mixup~\cite{verma2018manifold} & WRN & $46.21\pm0.77\%$ & $66.03\pm0.71\%$\\
         S2M2\_R~\cite{mangla2020charting} & WRN & $48.24\pm0.84\%$ & $70.44\pm0.75\%$\\
         Transfer+Graph Interpolation (ours) & WRN & $\mathbf{58.63\pm0.25\%}$ & $\mathbf{73.46\pm0.17\%}$\\
         \bottomrule
         
         \toprule
         &          & \multicolumn{2}{c}{\textbf{CIFAR-FS}} \\
         Method & Backbone & 1-shot & 5-shot \\
         \midrule
         BD-CSPN~\cite{liu2019prototype} & WRN & $72.13\pm1.01\%$ & $82.28\pm0.69\%$\\
         S2M2\_R~\cite{mangla2020charting} & WRN & $74.81\pm0.19\%$ & $87.47\pm0.13\%$\\
         SIB~\cite{hu2020empirical} & WRN & $80.00\pm0.60\%$ & $85.30\pm0.40\%$\\
         Transfer+Graph Interpolation (ours) & WRN & $\mathbf{83.90\pm0.22\%}$ & $\mathbf{88.76\pm0.15\%}$\\
         \bottomrule
         
    \end{tabular}
    }
    \label{tab:results}
\end{table*}

\subsection{Comparaison with other GNN methods}

In this experiment we compare our performance on miniImageNet with others that use Graph Neural Network to address the few-shot classification. As we can see in Table~\ref{tab:results_compareGNN}, with a three-stage training strategy, our proposed method has largely surpassed the current GNN based methods that train an entire model at once, given the transductive setting. 

\begin{table}
    \caption{1-shot and 5-shot performance (on miniImageNet) comparison with other GNN based methods. In our experiment we use the same hyperparameters as Table~\ref{tab:results}.}
    \centering
    \scalebox{0.9}{
    \begin{tabular}{l|l|l}
         \toprule
         Method & 1-shot & 5-shot \\
         \midrule
         GNN~\cite{garcia2017few} & $50.33\pm0.36\%$ & $66.41\pm0.63\%$\\
         TPN~\cite{liu2018learning} & $55.51\pm0.86\%$ & $69.86\pm0.65\%$\\
         wDAE-GNN~\cite{gidaris2019generating} & $61.07\pm0.15\%$ & $76.75\pm0.11\%$\\
         Transfer+Graph Interpolation (ours) & $\mathbf{76.50\pm0.23}\%$ & $\mathbf{85.23\pm0.13}\%$\\
         \bottomrule
         
    \end{tabular}
    }

    \label{tab:results_compareGNN}
\end{table}

\subsection{Importance of the parameter-free graph interpolation}

In our work, we considered using a parameter-free graph interpolation technique to diffuse features between inputs. As mentioned in the related work section, there are many alternatives, but they come with additional parameters. In the next experiment, we compare the accuracy of the method when using GCN~\cite{kipf2016semi} and GAT~\cite{velivckovic2017graph}, instead of a simple interpolation. Results are presented in Table~\ref{tab:results_compare}. We note that the best results are obtained using our designed graph interpolation, which we believe to be due to the fact we use fewer parameters in total. Graph interpolation also has the interest of being many times faster to train. In our experiments, each run took about $0.65$ seconds to train using graph interpolation versus $1.18$ seconds for GCN and $22.42$ seconds with GAT, which happens to lead to the worst performance of our considered methods.

It is worth pointing out that a drawback of the proposed method is that it requires to train a logistic regression model each time a batch prediction is required. In other words, it can be limiting in settings where predictions to make are streamed. However, the time required to train the logistic regression model remains very small in our experiments (less than one second).

\begin{table}
    \caption{1-shot and 5-shot accuracy on miniImageNet, when using the WRN backbone and various Graph Neural Networks. We use the same hyperparameters as Table~\ref{tab:results} and apply them to all methods (with the exception of $\kappa$ for GCN and GAT).}
    \centering
    \scalebox{1.0}{
    \begin{tabular}{l|l|l}
         \toprule
         Method & 1-shot & 5-shot \\
         \midrule
         Transfer+GAT & $65.38\pm0.89\%$ & $76.00\pm0.67\%$\\
         Transfer+GCN & $75.88\pm0.23\%$ & $84.51\pm0.13\%$\\
         Transfer+Graph Interpolation & $\mathbf{76.47\pm0.23\%}$ & $\mathbf{85.23\pm0.13\%}$\\
         \bottomrule
         \multicolumn{3}{l}{%
            \begin{minipage}{6.5cm}%
            \tiny *GAT is evaluated with 600 test runs.%
            \end{minipage}%
        }
    \end{tabular}
    }

    \label{tab:results_compare}
\end{table}

\subsection{Influence of Parameters}

\begin{figure}[h]
  \begin{center}
    \begin{tikzpicture}
       \begin{scope}[xscale=0.9, yshift=6cm]
        \begin{axis}[
            xlabel=$s$,
            ylabel=Accuracy,
            height=4cm,
            width=.5\textwidth,
            legend pos={south east,legend cell align=left}
            ]
          
          
          
          \addlegendentry{$Q/K_n=5$}
          \addplot coordinates
          {(1,66.70) (2,74.56) (3,78.19) (4,80.75) (5,82.18)};
          
          
          \addlegendentry{$Q/K_n=100$}
          \addplot coordinates
          {(1,79.62) (2,83.57) (3,85.20) (4,86.21) (5,86.95)};
        \end{axis}
      \end{scope}
    \end{tikzpicture}
  \end{center}
  \vspace{-.5cm}
  \caption{Evolution of the accuracy of few-shot classification with miniImageNet (backbone: WRN) as a function of the number of supervised inputs $s$, and for various number of unsupervised queries $q$. We use $\alpha=0.5$, $\kappa=3$ and $k=10$.}
  \label{fig:functionofs}
\end{figure}
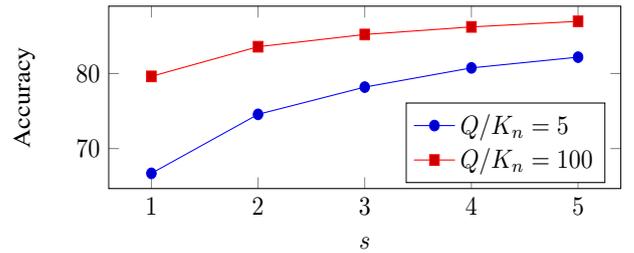

We then inquire the importance of various parameters of the task to the performance of the proposed method. We begin by varying the number of supervised inputs $s$, and consider two settings: one where we dispose of an average of $Q/K_n=5$ unsupervised inputs for each class and one where we dispose of $Q/K_n=100$ of them. Results are depicted in Figure~\ref{fig:functionofs}. As we can see, the performance of the method is highly influenced by the number of supervised inputs, as expected. Interestingly, there is a significant gap in accuracy between $Q/K_n=5$ and $Q/K_n=100$ for 1-shot setting, even if this gap diminishes as the number of supervised inputs is increased.

In the next experiment, we draw in Figure~\ref{fig:functionofq} the evolution of the performance of the method as a function of the number of unsupervised inputs $Q$, for 1-shot, 3-shot and 5-shot settings. This curve confirms two observations: a) in the case of 5-shot setting, the influence of the number of unsupervised inputs is little, and the accuracy of the method quickly reaches its pick and b) in the case of 1-shot setting, the number of unsupervised inputs significantly influences accuracy up to a few dozens. It is interesting to point out that about the same accuracy is achieved for 5-shot using $Q=1$ and 1-shot using $Q=100$, suggesting that $100$ unsupervised inputs bring about the same usable information as 4 labeled inputs per class.

\begin{figure}[h]
  \begin{center}
    \begin{tikzpicture}
       \begin{scope}[xscale=0.9, yshift=6cm]
        \begin{axis}[
            xlabel=$Q$,
            ylabel=Accuracy,
            height=4cm,
            width=.5\textwidth,
            legend pos=south east
            ]
            
          \addlegendentry{$s=1$}
          \addplot coordinates
          {(1,63.36) (25,66.72) (50,74.32) (75,76.33) (100,77.41) (250,79.00) (500,79.75)};
          
          \addlegendentry{$s=3$}
          \addplot coordinates
          {(1,70.68) (25,78.23) (50,81.55) (75,82.51) (100,83.16) (250,84.42) (500,85.38)};
          
          \addlegendentry{$s=5$}
          \addplot coordinates
          {(1, 77.00) (25,82.21) (50,84.02) (75,84.59) (100,85.15) (250,86.10) (500,86.93)};
        \end{axis}
      \end{scope}
    \end{tikzpicture}
  \end{center}
  \vspace{-.5cm}
  \caption{Evolution of the accuracy of few-shot classification with miniImageNet (backbone: WRN) as a function of the number of query inputs $Q$, and for various number of unsupervised inputs $s$. We use $\alpha=0.5$, $\kappa=3$ and $k=\min(10,s K_n + Q - 1)$.}
  \label{fig:functionofq}
\end{figure}

%
%
          %

\begin{figure}[h]
  \begin{center}
    \begin{tikzpicture}
       \begin{scope}[xscale=0.9, yshift=3cm]
        \begin{axis}[
            ylabel={\begin{tabular}{c}\% accuracy\\$k=10$\end{tabular}},
            height=4cm,
            width=.5\textwidth,
            ymin=60,
            ymax=77,
            xticklabels=[]
            ]
            
          \addplot coordinates
          {(1, 75.16) (2, 76.12) (3, 74.83) (4, 73.46) (5, 71.75)};

          \addplot coordinates
          {(1, 75.50) (2, 76.32) (3, 76.09) (4, 75.53) (5, 74.74)};
          
          \addplot coordinates
          {(1, 75.05) (2, 76.03) (3, 76.32) (4, 76.08) (5, 75.71)};
          
          \addplot coordinates
          {(1, 74.52) (2, 75.83) (3, 76.31) (4, 76.26) (5, 75.97)};
          
          \addplot coordinates
          {(1, 73.96) (2, 75.75) (3, 76.24) (4, 76.27) (5, 76.13)};
        \end{axis}
      \end{scope}
       \begin{scope}[xscale=0.9, yshift=5.7cm]
        \begin{axis}[
            legend style={at={(0.48,0.0)},anchor=south, legend cell align=left},
            ylabel={\begin{tabular}{c}\% accuracy\\$k=5$\end{tabular}},
            height=4cm,
            width=.5\textwidth,
            ymin=60,
            ymax=77,
            xticklabels=[]]
            
          \addlegendentry{\footnotesize{$\alpha=0$}}
          \addplot coordinates
          {(1, 73.71) (2, 75.67) (3, 75.14) (4, 75.17) (5, 74.63)};

          \addlegendentry{\footnotesize{$\alpha=0.25$}}
          \addplot coordinates
          {(1, 74.56) (2, 75.43) (3, 75.51) (4, 75.48) (5, 75.34) };
          
          \addlegendentry{\footnotesize{$\alpha=0.5$}}
          \addplot coordinates
          {(1, 74.55) (2, 75.41) (3, 75.59) (4, 75.52) (5, 75.39)};
          
          \addlegendentry{\footnotesize{$\alpha=0.75$}}
          \addplot coordinates
          {(1, 74.27) (2, 75.36) (3, 75.59) (4, 75.57) (5, 75.43)};
          
          \addlegendentry{\footnotesize{$\alpha=1$}}
          \addplot coordinates
          {(1, 73.85) (2, 75.29) (3, 75.59) (4, 75.60) (5, 75.49)};
        \end{axis}
      \end{scope}
       \begin{scope}[xscale=0.9, yshift=0.3cm]
        \begin{axis}[
            ylabel={\begin{tabular}{c}\% accuracy\\$k=15$\end{tabular}},
            height=4cm,
            width=.5\textwidth,
            ymin=60,
            ymax=77,
            xticklabels=[],
            legend pos=south east
            ]
            
          \addplot coordinates
          {(1, 74.13) (2, 74.14) (3, 71.44) (4, 68.78) (5, 66.30)};

          \addplot coordinates
          {(1, 74.92) (2, 75.24) (3, 74.29) (4, 72.84) (5, 71.25)};
          
          \addplot coordinates
          {(1, 74.40) (2, 75.37) (3, 75.20) (4, 74.54) (5, 73.65)};
          
          \addplot coordinates
          {(1, 73.74) (2, 75.20) (3, 75.35) (4, 75.13) (5, 74.60)};
          
          \addplot coordinates
          {(1, 73.14) (2, 74.90) (3, 75.32) (4, 75.27) (5, 75.01)};
        \end{axis}
      \end{scope}
       \begin{scope}[xscale=0.9, yshift=-2.4cm]
        \begin{axis}[
            xlabel=$\kappa$,
            ylabel={\begin{tabular}{c}\% accuracy\\$k=20$\end{tabular}},
            height=4cm,
            ymin=60,
            ymax=77,
            width=.5\textwidth,
            legend pos=south east
            ]
            
          \addplot coordinates
          {(1, 71.53) (2, 70.38) (3, 66.44) (4, 63.32) (5, 60.74)};

          \addplot coordinates
          {(1, 73.40) (2, 72.92) (3, 70.98) (4, 68.66) (5, 66.53)};
          
          \addplot coordinates
          {(1, 72.93) (2, 73.54) (3, 72.88) (4, 71.70) (5, 70.32)};
          
          \addplot coordinates
          {(1, 72.34) (2, 73.50) (3, 73.45) (4, 72.89) (5, 71.99)};
          
          \addplot coordinates
          {(1, 71.80) (2, 73.24) (3, 73.55) (4, 73.32) (5, 72.83)};
        \end{axis}
      \end{scope}
    \end{tikzpicture}
  \end{center}
  \vspace{-.5cm}
  \caption{Evolution of the accuracy of few-shot classification with miniImageNet (backbone: WRN) as a function of $\kappa$, $\alpha$ and $k$.}
  \label{fig:functionofkappa}
\end{figure}
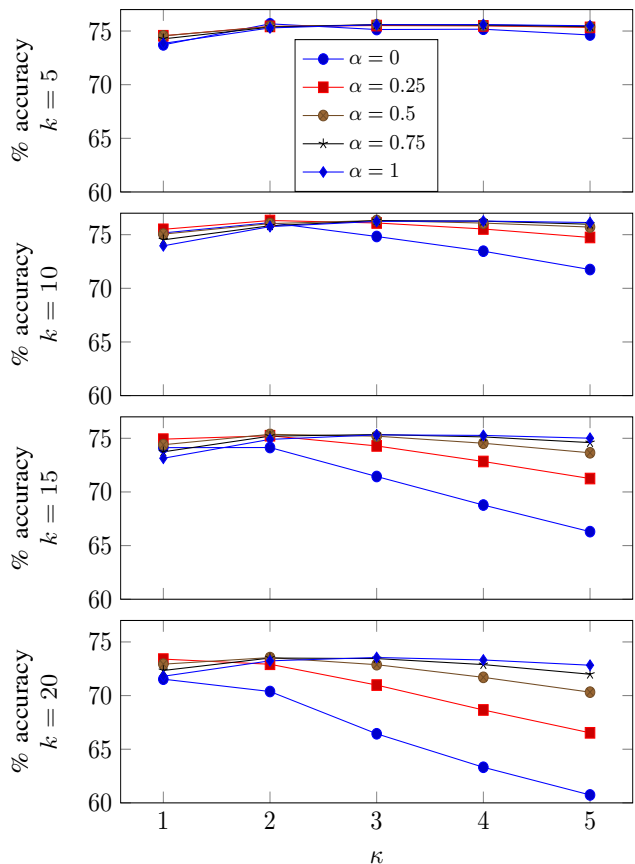

In the next experiment we look at the influence of the parameters $\kappa$ and $\alpha$ which respectively control to which power the diffusion matrix is taken and the importance of self-representations. In Figure~\ref{fig:functionofkappa}, we draw the obtained mean accuracy as a function of $\kappa$, $\alpha$ and $k$. We use $s=1$ and $Q/K_n = 15$ in this experiment. There are multiple interesting conclusions to draw from this figure. 

\begin{enumerate}
    \item This curve justifies the previously mentioned choice of parameters, leading to the best performance.
    \item We observe that when $k$ is large and $\alpha$ is small, it is better not to use powers of the diffusion matrix. This is the only setting where this statement holds, emphasizing the fact that if the graph is not sparse and self-importance is low, powers of the diffusion matrix are likely to over-smooth the representations of neighbors.
    \item When $k$ is small (here: $k=5$ or $k=10$), there is little sensitivity to both $\alpha$ and $\kappa$ (for $\kappa\leq 3$). This is an asset as it makes it simpler to find good hyperparameters.
    \item The best results are achieved for smaller values of $k$, suggesting that cosine similarity between distant representations can be noisy and damaging to the performance of the method.
    \item Note that in this experiment $s + Q/K_n = 16$. So using $k=15$ would ideally select exactly 15 neighbors of the same class for each input. Interestingly, this choice of $k$ does not lead to the best performance, showing the graph structure is not perfectly aligned with classes.
\end{enumerate}

It is often disregarded the impact of class imbalance in the context of few-shot learning. As a matter of fact, since we only consider very few labeled examples, it does not make much sense to consider such a scenario. But in the context of transductive setting, it is highly probable that unlabeled inputs are imbalanced between classes. So we perform the next experiment by varying the number of examples chosen in two random classes from miniImageNet. We always make sure that the total number of queries to classify remains the same, that is 100. But we select $q_1$ of them in class 1 and $100-q_1$ of them in class 2.

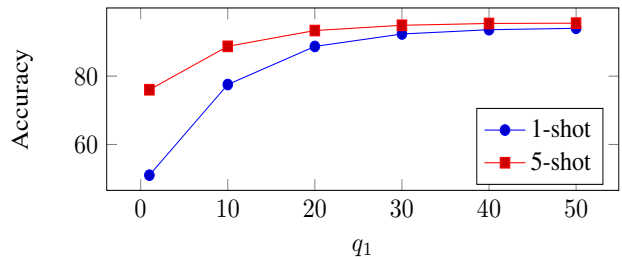
\begin{figure}[h]
  \begin{center}
    \begin{tikzpicture}
       \begin{scope}[xscale=0.9, yshift=6cm]
        \begin{axis}[
            xlabel=$q_1$,
            ylabel=Accuracy,
            height=4cm,
            width=.5\textwidth,
            legend pos=south east
            ]
            
          \addlegendentry{1-shot}
          \addplot coordinates
          {(1,51.02) (10,77.54) (20,88.69) (30,92.33) (40,93.58) (50,94.00)};
          
          \addlegendentry{5-shot}
          \addplot coordinates
          {(1,76.02) (10,88.72) (20,93.34) (30,94.88) (40,95.41) (50,95.50)};
        \end{axis}
      \end{scope}
    \end{tikzpicture}
  \end{center}
  \vspace{-.5cm}
  \caption{Accuracy of 2-ways classification with unevenly distributed query data for each class, where the total number of query inputs remains constant. When $q_1=1$, we obtain the most imbalanced case, whereas $q_1=50$ corresponds to a balanced case. We use $\alpha=0.5$, $\kappa=3$ and $k=10$.}
  \label{fig:imbalanced}
\end{figure}

In Figure~\ref{fig:imbalanced}, we depict the evolution of the accuracy of the proposed method, as a function of $q_1$. As one can clearly see from this figure, there is an important influence of class imbalance towards the performance of the proposed method. This is expected as the generated graphs will have imbalanced communities as a consequence. This could be problematic to some application domains where such imbalance is expected to happen in considered datasets, as there is no direct way of correcting it. Obviously, if one has insights about the relative distribution between classes, simple data augmentation or sampling could be used for mitigation.

However, this could be problematic to some application domains where such imbalance is expected to happen in considered datasets, as there is no direct way of correcting it. Obviously, if one has insights about the relative distribution between classes, simple data augmentation or sampling could be used for balancing this negative effect.

\begin{figure}[h]
\begin{center}
\begin{tikzpicture}[xscale=20,yscale=8]
\input{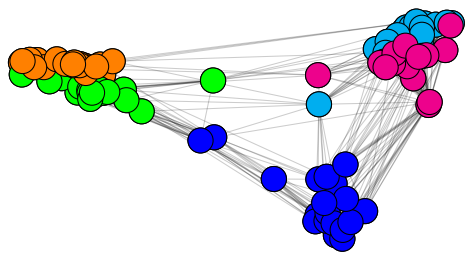}
\end{tikzpicture}
\end{center}
\caption{Visualisation of a graph obtained using miniImageNet. Colors represent various classes. Vertices are placed close if they share many connections.}
\label{graphvisu}
\end{figure}

Finally, in Figure~\ref{graphvisu}, we draw a representation of a typical graph obtained with the miniImageNet dataset, using Laplacian embedding~\cite{horaud2009short,shuman2013emerging}. On this figure, we colored vertices depending on which class they belong to. Interestingly, this figure shows that some classes are easily separated in the graph, whereas others are much harder to discriminate. We believe that the main reason why these graphs are not perfectly segregating classes is because some dimensions obtained using the backbone are specialized on features completely irrelevant for the novel task.

            %
          

\section{Conclusion}
\label{sec:con}
In this paper we introduced a novel method to solve the few-shot classification problem. It consists in combining three steps: a pretrained transfer, a graph-based interpolation technique and a logistic regression.

By performing experiments on standardized vision datasets, we obtained state-of-the-art results, with the most important gains in the case of 1-shot classification.

Interestingly, the proposed method requires to tune few hyperparameters, and these have a little impact on accuracy. We thus believe that it is an applicable solution to many practical problems.

There are still open questions to be addressed, such as the case of imbalanced classes, or settings where prediction must be performed on streaming data, one input at a time.






\bibliographystyle{IEEEtran}
\bibliography{IEEEabrv,egbib}
%



\end{document}